\documentclass[11pt]{article}
\usepackage[T1]{fontenc}
\usepackage{lmodern}
\usepackage[margin=1in]{geometry}
\usepackage{amsmath,amssymb}
\usepackage{booktabs}
\usepackage{graphicx}
\usepackage{xcolor}
\usepackage{microtype}
\usepackage{pgfplots}
\pgfplotsset{compat=1.18}
\usepackage[hidelinks]{hyperref}

\definecolor{gradblue}{RGB}{38,110,196}
\definecolor{randgrey}{RGB}{150,150,150}
\definecolor{refgrey}{RGB}{110,110,110}
\definecolor{stackteal}{RGB}{20,140,130}
\definecolor{stackorange}{RGB}{222,110,20}

\title{A Bias in Disguise: Tiny Gradient Edits Stop a Small\\Language Model's Thinking Loops and Complement\\a Frontier Model's Weight Edit}
\author{Hussamuddin Syed\\Independent Researcher\\\texttt{syed.hussam.uddin1@gmail.com}}
\date{November 2026}

\begin{document}
\maketitle

\begin{abstract}
Qwen3.5-0.8B rarely finishes reasoning in thinking mode: at a 4,096-token budget only 1 or 2 of 500 MATH-500 answers close their think block (two seeds), and accuracy is 15.4\%. A recent demonstration had GPT-6 Astra edit 131,072 of the model's weights without gradients and reported MATH-500 rising from 7.9\% to 48.1\% at a 32,768-token budget. We provide the missing gradient baseline, then combine the two edits. Fine-tuning only the 140 to 139,750 weights with the largest gradient, on the model's own correct answers to MATH training problems, raises thinking-mode MATH-500 at 4k tokens to 39.0--45.5\% after 2--9 minutes of training on one GPU; 50 training examples suffice, and 1,398 random positions give only 16.3\%. The gradient concentrates the edit on the most nearly constant channel of each of the first three layers. Because the model has no bias terms, an edit confined to those channels acts as a constant vector added to the residual stream: adding three fixed vectors to the unedited model matches the accuracy of such an edit. Under the Astra protocol (32k tokens, presence penalty 1.5), the Astra edit scores 52.1\% (finished and correct) in our reproduction and our full-strength edits reach 45.6--45.7\%, but ours win at every budget up to 16k tokens (39.5\% against 6.7\% at 4k). The two edits are complementary. The Astra edit plus our 139,750-weight edit at half strength scores 62.2\% at 32k, 10 points above the Astra edit alone ($p=3\times10^{-9}$), with half as many tokens; our half-strength edit alone scores 46.7\%.
\end{abstract}

\section{Introduction}

Small reasoning models often loop: they repeat ``Wait, let me reconsider'' until the token budget runs out. Pipis et al.~\cite{pipis2026loop} trace looping to errors in learning, and Liquid AI~\cite{liquid2026antidoom} report it in 22.9\% of greedy Qwen3.5-4B completions. In Qwen3.5-0.8B~\cite{qwen35} with thinking on, it is almost universal.

\emph{Sparks of Weight-Space Programming}~\cite{takehi2026sparks} let GPT-6 Astra read and edit this model's weights in a Python session, with no gradients. Astra changed 64 columns of \texttt{mlp.down\_proj} in layers 15 and 19: 131,072 weights, 0.015\% of the model's 873M. MATH-500 rose from 7.9\% to 48.1\%, and answers hitting the 32,768-token limit fell from 93\% to 33\%. The authors write that the edit ``lets the model finish answers it could already work out''. A follow-up~\cite{hu2026exploring} reports a 29.45-point MATH-500 gain from an edit of 1,023 parameters.

Neither work compares against ordinary gradient training of an equally small set of weights, or against randomly placed edits. Without that baseline, it is unclear what a frontier model adds. We ask four questions: how small a gradient-found edit that stops the looping can be, and how much data it needs; what the edit is, mechanically; how it compares with the Astra edit under each edit's own protocol; and whether the two edits can be combined.

Our contributions:
\begin{itemize}
\item \textbf{A cheap baseline.} Sparse fine-tuning on the model's own answers finds edits of 140 to 139,750 weights that take thinking-mode MATH-500 from 15.4\% to 39.0--45.5\%, in 2--9 minutes on one GPU.
\item \textbf{Controls.} At 1,398 weights, random positions fail (16.3\%) while gradient-chosen ones work (42.9\%). At 139,750 weights, about the size of the Astra edit, even random positions work (43.4\%), so an edit of that size is not small for this behaviour.
\item \textbf{A mechanism.} The gradient concentrates the edit on the most nearly constant channel of layers 0, 1 and 2. In a model without bias terms, an edit to those channels is a constant vector in disguise: for an edit trained only there, adding three fixed vectors matches its accuracy.
\item \textbf{A comparison.} The Astra edit lives in other layers, points in a nearly orthogonal direction and depends on the input. Under its own 32k-token protocol it beats our edits (52.1\% against 45.7\%), but ours win at every budget up to 16k: our edit stops early, theirs needs room to think.
\item \textbf{A combination.} Adding our edit at half strength to the Astra edit gives 62.2\% at 32k, above either edit alone (52.1\% and 46.7\%), with half the tokens of the Astra edit.
\end{itemize}

\section{Setup}
\label{sec:setup}

\paragraph{Model.} Qwen3.5-0.8B (Apache-2.0) has 873M parameters, 752M of them in the language model. Its 24 layers repeat three linear-attention layers and one full-attention layer. The hidden size is 1,024, and no linear layer has a bias term. We report edit sizes as a share of all 873M weights.

\paragraph{Training data.} With thinking off, the base model answered each of the 7,498 MATH~\cite{hendrycks2021math} training problems four times (temperature 0.7, top-$p$ 0.8, top-$k$ 20, 1,536 tokens). We kept finished answers graded correct, at most two per problem: 7,111 examples. Each was rewritten in thinking format. The prompt ends with \texttt{<think>}, and the target is the same reasoning, then \texttt{</think>} and the line ``The final answer is \texttt{\textbackslash boxed\{X\}}.'' Loss is computed on target tokens only. No test problem and no other model is involved.

\paragraph{Choosing the weights.} We run 64 micro-batches (about 4,096 tokens each) of this data through the frozen model. For every weight of the 150 projection matrices (attention, linear attention and MLP, in all 24 layers), we sum the absolute gradient. The $k$ weights with the largest sums, ranked across the whole model, become trainable. This is a gradient-magnitude variant of the FISH mask~\cite{sung2021fish}. The random control ranks uniform noise instead. The column variants in Section~\ref{sec:where} train every entry of chosen weight columns.

\paragraph{Training.} Only the $k$ chosen values change. We use AdamW at learning rate $3\times10^{-4}$, 20 warmup steps and cosine decay to 10\%, for one epoch: 361 steps of about 16k tokens. The 500- and 50-example runs repeat their data for 8 and 40 epochs. The edit is added to the original weights in fp32 and rounded to bf16. We count edit sizes as edited positions, $k$ for our edits and 131,072 for the Astra edit; after rounding, slightly fewer values change (1,397 of 1,398; 138,772 of 139,750; 130,188 of 131,072).

\paragraph{Evaluation.} We use vLLM~\cite{kwon2023vllm} on one A100 GPU (40\,GB). MATH-500~\cite{lightman2023verify} is graded with math-verify~\cite{mathverify} on the last boxed answer. Thinking-on runs sample at temperature 0.6, top-$p$ 0.95, top-$k$ 20, with no presence penalty and a 4,096-token budget (7,600 for ``8k'', so that prompt and answer fit the 8,192-token context we set; 3 of 500 MATH-500 prompts are long enough to cut it slightly). Each thinking-on setting is run with two seeds. Thinking-off runs use greedy decoding with 2,048 tokens. We also run GSM8K~\cite{cobbe2021gsm8k} (1,319 problems), HumanEval~\cite{chen2021humaneval} (164) and LiveCodeBench v6~\cite{jain2024livecodebench} (175). An answer is \emph{closed} if it finished and contains \texttt{</think>}. Comparisons on the same problems use an exact McNemar test~\cite{mcnemar1947} on the discordant pairs; with two samples per problem, we use a sign test on problems where the share of correct samples differs.

\paragraph{The Astra protocol.} Section~\ref{sec:protocol} follows the Astra evaluation: vLLM 0.33.0 (their version) on one A100 (80\,GB), up to 32,768 new tokens, temperature 0.6, top-$p$ 0.95, top-$k$ 20, presence penalty 1.5, two samples per problem (they use four). We report \emph{strict} accuracy: finished, closed and correct. Accuracy at a smaller budget $B$ is read off the same answers: an answer counts if it finished within $B$ tokens. A token limit does not change the tokens generated before it, so this has the same distribution as a run at budget $B$; direct runs at 4k, 8k and 16k for two models agree with it within 1.1 points. Combined edits add our edit, optionally scaled by a strength $s$, to the Astra checkpoint's weights; only 58 of our 139,750 positions are among the 130,188 weights in which the Astra checkpoint differs from the base model.

\section{Results}

\begin{figure}[t]
\centering
\begin{tikzpicture}
\begin{axis}[
  width=\linewidth, height=7.2cm,
  xmode=log, xmin=60, xmax=300000,
  ymin=0, ymax=52,
  xlabel={Weights edited (log scale)},
  ylabel={MATH-500, thinking on, 4k (\%)},
  xtick={100,1000,10000,100000},
  xticklabels={100,1{,}000,10{,}000,100{,}000},
  ytick={0,10,20,30,40,50},
  grid=major, grid style={gray!20},
  axis lines=left,
  legend style={at={(0.98,0.03)}, anchor=south east, draw=none, font=\small, fill=white, fill opacity=0.9, text opacity=1},
  legend cell align=left,
  tick label style={font=\small}, label style={font=\small},
]
\addplot[refgrey, dashed, thick, domain=60:300000, samples=2, forget plot] {41.0};
\node[anchor=south west, font=\footnotesize, text=refgrey] at (axis cs:65,41.2) {Base, thinking off: 41.0\%};
\addplot[refgrey, dashed, thick, domain=60:300000, samples=2, forget plot] {15.4};
\node[anchor=south west, font=\footnotesize, text=refgrey] at (axis cs:65,15.6) {Base, thinking on: 15.4\%};
\addplot[gradblue, very thick, mark=*, mark size=2.5pt,
  error bars/.cd, y dir=both, y explicit, error bar style={gradblue, thick}]
  coordinates {(140,39.0)+-(0,0.6) (500,40.3)+-(0,0.3) (1398,42.9)+-(0,1.9) (13975,44.0)+-(0,0.2) (139750,45.5)+-(0,1.3)};
\addlegendentry{Gradient-chosen weights}
\addplot[randgrey, only marks, mark=*, mark size=2.5pt,
  error bars/.cd, y dir=both, y explicit, error bar style={randgrey, thick, solid}]
  coordinates {(1398,16.3)+-(0,0.9) (139750,43.4)+-(0,0.2)};
\addlegendentry{Random positions}
\addplot[black!70, only marks, mark=diamond*, mark size=3.5pt,
  error bars/.cd, y dir=both, y explicit, error bar style={black!70, thick}]
  coordinates {(131072,18.9)+-(0,0.3)};
\addlegendentry{GPT-6 Astra edit (our protocol)}
\node[anchor=north, font=\footnotesize\bfseries, text=gradblue] at (axis cs:140,37.6) {39.0\%};
\node[anchor=south, font=\footnotesize\bfseries, text=gradblue] at (axis cs:1398,45.2) {42.9\%};
\node[anchor=north, font=\footnotesize, text=randgrey!80!black] at (axis cs:1398,14.2) {16.3\%};
\end{axis}
\end{tikzpicture}
\caption{Thinking-mode MATH-500 accuracy against edit size. Points are the mean of two samples; bars span both. 140 gradient-chosen weights reach 39.0\%. Random positions fail at 1,398 weights and work at 139,750; we did not test sizes in between. The Astra edit (131,072 weights) is evaluated with our 4,096-token protocol, not its own 32,768-token one.}
\label{fig:size}
\end{figure}
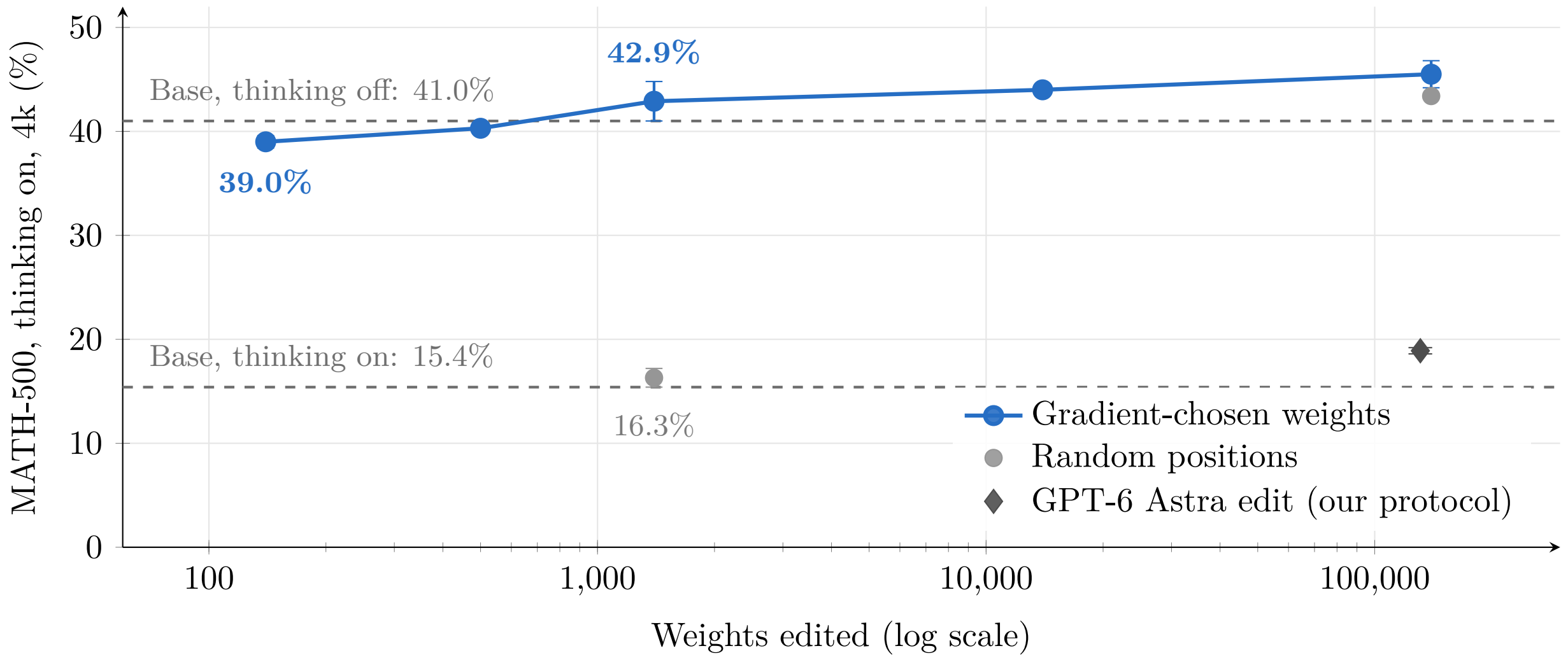

A 140-weight gradient edit captures most of the fix. Random positions fail at 1,398 weights and match gradient-chosen ones only at 139,750, the largest size we tried (Figure~\ref{fig:size}). At 140 weights, a learning rate of $10^{-3}$ gave the same result (38.4\% and 37.4\%), so step size does not limit the smallest edit.

\paragraph{Data needed.} Fifty examples are enough (Table~\ref{tab:data}). The 50- and 500-example sets were drawn from the full pool, which took 57.6 A100-minutes to generate; generating 50 examples directly would take a small fraction of that.

\begin{table}[!htbp]
\centering
\small
\begin{tabular}{rrrcc}
\toprule
Training examples & Epochs & Training (min) & MATH-500, thinking 4k & Closed (of 500) \\
\midrule
7,111 & 1 & 6.8 & 41.0\% / 44.8\% & 344 / 340 \\
500 & 8 & 3.7 & 44.8\% / 43.6\% & 351 / 351 \\
50 & 40 & 1.9 & 41.8\% / 37.4\% & 361 / 356 \\
\bottomrule
\end{tabular}
\caption{Amount of training data, all at 1,398 weights. Two samples per setting.}
\label{tab:data}
\end{table}

\paragraph{Against thinking off.} The fair baseline is the base model with thinking off, which scores 41.0\% (greedy decoding, while the thinking-on runs are sampled). The 1,398-weight edit matches it: 41.0\% ($p=1.0$) and 44.8\% ($p=0.10$). The 139,750-weight edit beats it at the 8k budget, 48.0\% ($p=0.002$), and scores 46.8\% ($p=0.009$) and 44.2\% ($p=0.18$) at 4k. With thinking off, the edited models score 40.2--41.4\%, the same as the base model. The edits make thinking mode usable but add little reasoning ability. A dense LoRA (rank 16, changing 409M weights) reaches 41.0\%, no better than 1,398 sparse weights.

\paragraph{Transfer.} The fix carries over to GSM8K, which it never trained on, but not to code (Table~\ref{tab:transfer}). On HumanEval, the edit closes fewer thinking blocks than the base model does, and with thinking off it scores 29.3\% against 34.1\%: it may slightly hurt code, though no code difference is significant.

\begin{table}[!htbp]
\centering
\small
\resizebox{\linewidth}{!}{%
\begin{tabular}{lccc}
\toprule
Benchmark & Base & 139,750-weight edit & Fixed / broken vs base \\
\midrule
GSM8K, thinking 4k & 13.6\% (18 of 1,319 closed) & 56.1\% (1,244 closed) & +605 / $-$44, $p<10^{-100}$ \\
HumanEval, thinking 4k & 27.4\% (61 of 164 closed) & 23.8\% (36 closed) & +14 / $-$20, $p=0.39$ \\
GSM8K, thinking off & 55.5\% & 54.9\% & $p=0.69$ \\
HumanEval, thinking off & 34.1\% & 29.3\% & $p=0.15$ \\
LiveCodeBench v6, thinking off & 9.1\% & 10.3\% & $p=0.62$ \\
\bottomrule
\end{tabular}}
\caption{Transfer to benchmarks outside the training data.}
\label{tab:transfer}
\end{table}

\section{Where the fix lives}
\label{sec:where}

The gradient concentrates the edit on the most nearly constant channel of each of the first three layers, and an edit confined to those channels acts as a constant vector added to the residual stream. The rest of the edit matters too, and we have not tested whether it acts the same way.

\paragraph{Four constant channels.} In the 1,398-weight edit, 888 weights (64\%) fall in four input columns of \texttt{linear\_attn.out\_proj} in layers 0--2. Each column is how one channel of one value head writes into the residual stream. They are the most constant channels in their layers, measured on 19 thinking traces with the base weights (Table~\ref{tab:channels}). The smaller edits use the same channels: 113 of the 140 weights and 380 of the 500 sit in these four columns.

\begin{table}[!htbp]
\centering
\small
\begin{tabular}{lrrrrr}
\toprule
Channel & Mean & Std & Mean/std (rank of 2,048) & Layer median & Weights in edit \\
\midrule
Layer 0, channel 1105 & +2.244 & 0.176 & 12.8 (1st) & 0.24 & 317 \\
Layer 0, channel 826 & $-$2.126 & 0.445 & 4.8 (2nd) & 0.24 & 135 \\
Layer 1, channel 1583 & $-$2.198 & 0.354 & 6.2 (1st) & 0.11 & 245 \\
Layer 2, channel 1492 & $-$2.688 & 0.067 & 40.2 (1st) & 0.15 & 191 \\
\bottomrule
\end{tabular}
\caption{The four channels the 1,398-weight edit concentrates on: activation statistics under the base weights. ``Layer median'' is the median $|\text{mean}|/\text{std}$ over the layer's 2,048 channels.}
\label{tab:channels}
\end{table}

\paragraph{Ablations and column training.} Training only these four columns works as well as the full edit, and four random columns do not (Table~\ref{tab:ablation}). The parts of the 1,398-weight edit do not add up: each alone closes about 65 answers, together 342. The push seems to work only above a threshold. The four channels are not the whole edit: its 510 weights outside them score higher alone (about 33\%) than the 888 inside them (about 25\%).

\begin{table}[!htbp]
\centering
\small
\begin{tabular}{lrcc}
\toprule
Variant & Weights & MATH-500, thinking 4k & Closed (of 500) \\
\midrule
Full 1,398-weight edit & 1,398 & 41.0\% / 44.8\% & 344 / 340 \\
Only its part in the four channels & 888 & 25.0\% / 24.8\% & 62 / 59 \\
Only its other part & 510 & 34.6\% / 30.8\% & 67 / 64 \\
Only its part in channel 1105 & 317 & 16.6\% / 17.0\% & 13 / 11 \\
Train the four channel columns from scratch & 4,096 & 45.8\% / 44.0\% & 347 / 335 \\
Train four random columns, same matrices & 4,096 & 20.4\% / 20.6\% & 24 / 18 \\
All three vectors packed into one column & 1,024 & 21.2\% / 20.4\% & 41 / 29 \\
\bottomrule
\end{tabular}
\caption{Ablations of the 1,398-weight edit and column-only training. Two samples per variant.}
\label{tab:ablation}
\end{table}

\paragraph{A bias in disguise.} Qwen3.5's linear layers have no bias terms. If channel $c$ is nearly constant at mean $\mu_c$, changing its write column by $\Delta_{:,c}$ adds almost the same vector at every position. For layer $L$ with edited channels $C_L$:
\begin{equation}
b_L = \sum_{c \in C_L} \mu_c\, \Delta_{:,c}.
\end{equation}
We tested this on the four-column edit. Each $b_L$ has norm 0.23--0.28. Hooks add $b_0$, $b_1$ and $b_2$ to the unedited model's \texttt{out\_proj} outputs in layers 0--2. On 38 teacher-forced thinking traces, the mean KL divergence from the edited model is 0.010 nats per token for base + vectors, against 0.108 for the base model, and the top-1 token agrees 98.1\% of the time, against 93.2\%. On the first 96 MATH-500 problems (Hugging Face sampling, same settings), base + vectors scores 45.8\% with 57 answers closed; the edited model scores 42.7\% (69 closed) and the base model 11.5\% (1 closed). The vectors match the four-column edit's accuracy, but not all of its closing.

This explains why the gradient picks these channels. The training signal asks for the same shift everywhere: finish sooner. Without biases, the cheapest way to add a constant is through a channel that is already constant. Placement matters, though: writing $b_0+b_1+b_2$ into the layer-0 column alone fails (20.8\%), although per-layer placement works.

\paragraph{What the push changes.} The edit does not simply raise the probability of \texttt{</think>} everywhere. On the base model's looping traces, the edited model's mean log-probability of \texttt{</think>} is lower than the base model's ($-29.7$ against $-26.9$), but its highest value per trace is higher ($-5.8$ against $-8.0$). On the edited model's own finished traces, the base model would also close at the same position: median log-probability $-0.13$, above probability 0.5 in 94\% of traces. The base model recognises when it is done. What it fails to do is reach that point, and the edit changes the path taken before it.

\section{Comparison with the GPT-6 Astra edit}
\label{sec:astra}

The Astra edit is a different, input-dependent change in the middle of the network. At our 4k--8k budgets it rarely ends the loops, and stacking it on our edit adds nothing there. Section~\ref{sec:protocol} shows that this changes at their 32k budget.

\paragraph{What their diff is.} Only two tensors of the public checkpoint~\cite{astra2026ckpt} differ from the base model: \texttt{mlp.down\_proj} in layers 15 and 19. In each, 64 columns change, and the change has rank one: one direction $u$ carries all of it (fit residual 0.2--0.3\%). On those columns,
\begin{equation}
W' = W - 6\, u u^{\top} W .
\end{equation}
This multiplies the component along $u$ that these 64 neurons write by $-5$: the signal is flipped and amplified, not removed. Both layers use the same direction up to sign ($|\cos| = 0.85$).

\paragraph{How it differs from ours.} Table~\ref{tab:diff} summarises the two edits. They point in nearly orthogonal directions. The vectors written by the four columns of our 1,398-weight edit have $|\cos|$ with $u$ of at most 0.02, 0.03 and 0.05 for three columns. The fourth, in layer 2, reaches 0.10, above the 0.08 that random directions reach at the 99th percentile, so the overlap is small but not zero: 1\% of that vector's squared length lies along $u$. Their edit tracks the loops. The amount the 64 neurons write along $u$ varies with position (mean/std 0.93 in layer 15, 0.76 in layer 19), and is 40--56\% larger on looping traces than on finished ones (0.098 against 0.063 in layer 15; 0.129 against 0.091 in layer 19). Our edit does not act through their direction: on looping traces it leaves the residual stream's cosine with $u$ almost unchanged (0.206 against 0.214 at layer 15, 0.251 against 0.244 at layer 19), while the Astra edit lowers it to 0.065 and 0.099.

\begin{table}[!htbp]
\centering
\small
\begin{tabular}{lp{5.6cm}p{5.6cm}}
\toprule
 & Our four-column edit & Astra edit \\
\midrule
Found by & Gradient descent on 7,111 of the model's own answers, 9 min on one A100 & GPT-6 Astra reading the weights, with 50 forward passes and no gradients \\
Where & Layers 0--2, \texttt{linear\_attn.out\_proj} & Layers 15 and 19, \texttt{mlp.down\_proj} \\
Weights & 4,096 (140 in our smallest edit) & 131,072 \\
What it adds & A near-constant vector & An amount that depends on the input \\
\bottomrule
\end{tabular}
\caption{The two edits side by side.}
\label{tab:diff}
\end{table}

\paragraph{Their checkpoint on our protocol.} Table~\ref{tab:astra} runs the public Astra checkpoint through our evaluation. Stacking the Astra edit on our 1,398-weight edit gives 43.8\% and 40.4\%, against 41.0\% and 44.8\% for ours alone. Their edit does close more code answers than the base model (105 against 61), where ours closes fewer.

\begin{table}[!htbp]
\centering
\small
\resizebox{\linewidth}{!}{%
\begin{tabular}{lccc}
\toprule
Benchmark & Base & Astra edit & Ours, 139,750 weights \\
\midrule
MATH-500, thinking 4k & 16.0\% / 14.8\% (1 / 2 closed) & 18.6\% / 19.2\% (14 / 15 closed) & 46.8\% / 44.2\% (349 / 363 closed) \\
MATH-500, thinking 8k & 15.6\% (3 closed) & 24.0\% (45 closed) & 48.0\% (375 closed) \\
GSM8K, thinking 4k & 13.6\% (18 of 1,319 closed) & 21.5\% (120 closed) & 56.1\% (1,244 closed) \\
HumanEval, thinking 4k & 27.4\% (61 of 164 closed) & 32.3\% (105 closed) & 23.8\% (36 closed) \\
MATH-500, thinking off & 41.0\% & 37.2\% ($p=0.07$) & 40.4\% \\
GSM8K, thinking off & 55.5\% & 57.3\% & 54.9\% \\
HumanEval, thinking off & 34.1\% & 36.6\% & 29.3\% \\
\bottomrule
\end{tabular}}
\caption{The public Astra checkpoint and our 139,750-weight edit under our protocol (no presence penalty, 4k or 8k tokens).}
\label{tab:astra}
\end{table}

This is not their protocol, which uses 32,768 tokens and presence penalty 1.5. We run it next.

\section{Under the Astra protocol: each edit wins in its own regime, and together they win}
\label{sec:protocol}

\begin{table}[!htbp]
\centering
\small
\resizebox{\linewidth}{!}{%
\begin{tabular}{lrcccrl}
\toprule
Model & Weights edited & Strict & Any correct & Hit the limit & Avg tokens & vs Astra edit (strict) \\
\midrule
Base (Astra report) & 0 & -- & 7.9\% & 93\% & -- & \\
Astra edit (Astra report / re-run) & 131,072 & -- & 48.1 / 51.0\% & 33.1 / 31.6\% & -- & \\
Astra edit (ours) & 131,072 & 52.1\% & 55.7\% & 32.0\% & 21,966 & \\
\midrule
Ours, 1,398 & 1,398 & 45.7\% & 47.1\% & 11.5\% & 6,861 & $-$6.4, $p=2\times10^{-6}$ \\
Ours, 139,750 & 139,750 & 45.6\% & 46.1\% & 16.2\% & 7,501 & $-$6.5, $p=10^{-7}$ \\
Ours, 139,750 at half strength & 139,750 & 46.7\% & 50.8\% & 42.9\% & 16,230 & $-$5.4, $p=6\times10^{-4}$ \\
\midrule
Astra + ours, 1,398 & 132,470 & 47.6\% & 48.8\% & 0.8\% & 5,188 & $-$4.5, $p=0.03$ \\
Astra + ours, 139,750 & 270,764 & 49.7\% & 49.7\% & 0.5\% & 4,382 & $-$2.4, $p=0.11$ \\
\textbf{Astra + ours, 139,750 at half strength} & 270,764 & \textbf{62.2\%} & \textbf{62.9\%} & 8.2\% & 11,523 & \textbf{+10.1}, $p=3\times10^{-9}$ \\
Astra + ours, 139,750 at quarter strength & 270,764 & 59.8\% & 61.5\% & 20.9\% & 17,483 & +7.7, $p=10^{-8}$ \\
\bottomrule
\end{tabular}}
\caption{MATH-500 under the Astra protocol (32,768 tokens, presence penalty 1.5). Two samples per problem for our runs; the Astra report uses four. \emph{Strict}: finished, closed and correct. \emph{Any correct}: the last boxed answer is correct, finished or not. We list the Astra report's accuracy under \emph{any correct}. \emph{Weights edited} counts positions; the 139,750-weight edit and the Astra edit share 58, so together they edit $131{,}072 + 139{,}750 - 58 = 270{,}764$.}
\label{tab:protocol}
\end{table}

\paragraph{Reproduction.} Our run of the public Astra checkpoint scores 55.7\% (any correct), with 32.0\% of answers hitting the limit, against 51.0\% and 31.6\% in their re-run (Table~\ref{tab:protocol}). Its strict score is 52.1\%. Under this protocol the Astra edit beats each of our edits alone, by 5--6 points.

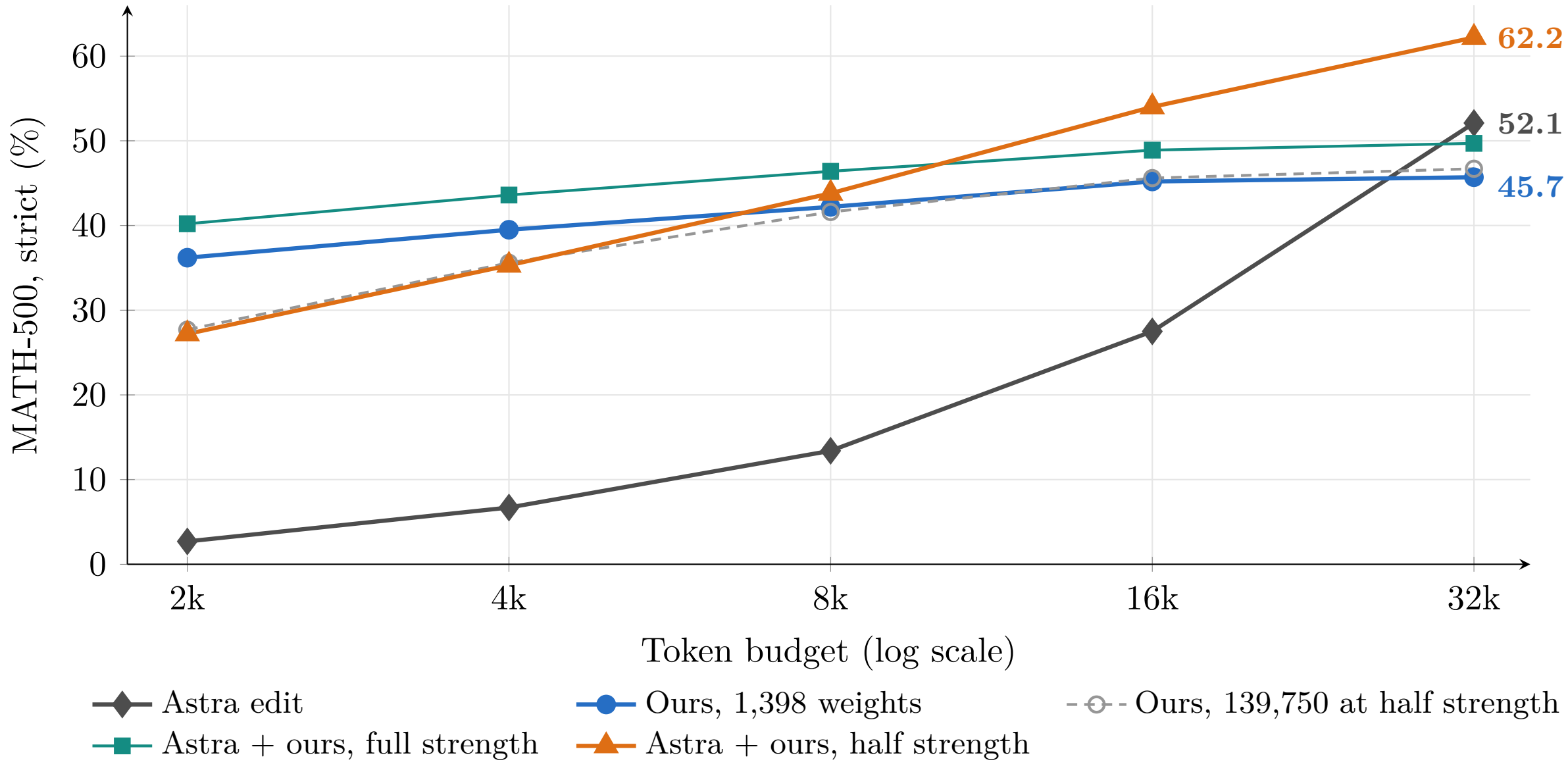
\begin{figure}[t]
\centering
\begin{tikzpicture}
\begin{axis}[
  width=0.95\linewidth, height=7.2cm,
  xmode=log, log basis x=2, xmin=1800, xmax=37000, clip=false,
  ymin=0, ymax=66,
  xlabel={Token budget (log scale)},
  ylabel={MATH-500, strict (\%)},
  xtick={2048,4096,8192,16384,32768},
  xticklabels={2k,4k,8k,16k,32k},
  ytick={0,10,20,30,40,50,60},
  grid=major, grid style={gray!20},
  axis lines=left,
  legend style={at={(0.5,-0.2)}, anchor=north, draw=none, font=\footnotesize, legend columns=3, /tikz/every even column/.append style={column sep=8pt}},
  legend cell align=left,
  tick label style={font=\small}, label style={font=\small},
]
\addplot[black!70, very thick, mark=diamond*, mark size=3pt] coordinates {(2048,2.7) (4096,6.7) (8192,13.4) (16384,27.5) (32768,52.1)};
\addlegendentry{Astra edit}
\addplot[gradblue, very thick, mark=*, mark size=2.2pt] coordinates {(2048,36.2) (4096,39.5) (8192,42.2) (16384,45.2) (32768,45.7)};
\addlegendentry{Ours, 1,398 weights}
\addplot[randgrey, thick, densely dashed, mark=o, mark size=2.2pt, mark options={solid}] coordinates {(2048,27.7) (4096,35.6) (8192,41.6) (16384,45.6) (32768,46.7)};
\addlegendentry{Ours, 139,750 at half strength}
\addplot[stackteal, thick, mark=square*, mark size=2pt] coordinates {(2048,40.2) (4096,43.6) (8192,46.4) (16384,48.9) (32768,49.7)};
\addlegendentry{Astra + ours, full strength}
\addplot[stackorange, very thick, mark=triangle*, mark size=3pt] coordinates {(2048,27.2) (4096,35.3) (8192,43.8) (16384,54.0) (32768,62.2)};
\addlegendentry{Astra + ours, half strength}
\node[anchor=west, font=\footnotesize\bfseries, text=stackorange] at (axis cs:33500,62.2) {62.2};
\node[anchor=west, font=\footnotesize\bfseries, text=black!70] at (axis cs:33500,52.1) {52.1};
\node[anchor=west, font=\footnotesize\bfseries, text=gradblue] at (axis cs:33500,44.6) {45.7};
\end{axis}
\end{tikzpicture}
\caption{Strict MATH-500 accuracy against token budget, Astra protocol. Points below 32k are read off the 32k answers (Section~\ref{sec:setup}). Our edit is close to its best by 4k; the Astra edit needs about 20k tokens per answer to pass it. The combined edit is best at 16k and 32k; with our edit at full strength it is best at 8k and below.}
\label{fig:budget}
\end{figure}

\paragraph{Each edit wins in its own regime.} Our edits finish early: half of our 1,398-weight edit's answers use under 1,200 tokens, and its score at 4k (39.5\%) is within 7 points of its score at 32k. The Astra edit thinks long. Its median answer uses 23,000 tokens, and it scores 6.7\% at 4k and 27.5\% at 16k. Ours is better at every budget up to 16k; the Astra edit passes it between 16k and 32k (Figure~\ref{fig:budget}). The two edits fail in opposite ways. When the Astra edit closes its thinking, it is right 76.8\% of the time, but only 67.8\% of its answers close; most of the rest run out of tokens. Our 1,398-weight edit closes 87.6\% of its answers but is right on only 52.2\% of those.

\paragraph{Together they win.} Adding our 139,750-weight edit to the Astra edit at full strength fixes the running out (0.5\% hit the limit) but undoes the long thinking: the median answer falls to 1,068 tokens, and the score to 49.7\%. At half strength, the combined model keeps most of the long thinking and loses most of the running out: 91.8\% of answers close, and 67.8\% of those are right. It scores 62.2\% at 32k, 10.1 points above the Astra edit (better on 135 problems, worse on 54, $p=3\times10^{-9}$), with 11,523 tokens per answer against 21,966. Our half-strength edit on its own scores 46.7\% and runs out of tokens 42.9\% of the time, so the gain needs both edits ($p=2\times10^{-19}$ against it). The combination is also best at 16k (54.0\% against 27.5\% for the Astra edit); at 8k and below, the full-strength combination is best (43.6\% at 4k). One plausible reading: the Astra edit improves how the model reasons, and a weakened version of our constant push makes it stop once it is done.

\section{Related work}

\paragraph{Weight-space programming.} Takehi et al.~\cite{takehi2026sparks} have a frontier model read and rewrite network weights without gradients, and stop Qwen3.5-0.8B's loops with a 131,072-weight edit. Hu et al.~\cite{hu2026exploring} study the approach across 16 editing runs and two model sizes, including a 1,023-parameter edit worth 29.45 MATH-500 points. Neither compares against gradient-chosen sparse edits or random-position controls, which is our focus, and neither combines a model-written edit with a gradient edit.

\paragraph{Editing and steering reasoning length.} ThinkEdit~\cite{sun2025thinkedit} finds a linear direction for reasoning length and edits the output projections of about 4\% of attention heads (0.2\% of parameters) to fix reasoning that is too short. Sheng et al.~\cite{sheng2025planning} find a direction, set before generation, whose strength controls reasoning length and shifts the probability of \texttt{</think>}. Our edit addresses the opposite problem, reasoning that never ends. It is found by gradient descent rather than by contrasting activations, and its four-column core amounts to a constant steering vector stored in the weights.

\paragraph{Why models loop.} Pipis et al.~\cite{pipis2026loop} attribute looping to errors in learning, which higher temperature only partly hides. Final Token Preference Optimization~\cite{liquid2026antidoom} trains against the first token of a loop, cutting greedy looping in Qwen3.5-4B from 22.9\% to 1\%.

\paragraph{Sparse fine-tuning.} Choosing the $k$ parameters with the largest importance score and training only those is the FISH mask~\cite{sung2021fish}, which uses Fisher information; we use summed absolute gradients. Mukherjee et al.~\cite{mukherjee2025subnetworks} show that reinforcement learning updates only 5--30\% of an LLM's parameters, and that training that subnetwork alone recovers the result.

\paragraph{Large gains from small signals.} Shao et al.~\cite{shao2025spurious} report that reinforcement learning with random rewards raises Qwen2.5-Math-7B on MATH-500 by 21.4 points, close to the 29.1 from true rewards, by amplifying behaviour learned in pretraining. Our result is another large benchmark jump that unlocks existing ability rather than adding new ability.

\section{Limitations}

All results are for one model, Qwen3.5-0.8B, trained on MATH problems; we have not tested other sizes, other model families, or models with bias terms, where the constant-channel route may not be needed. Thinking-on results use two seeds, and dense LoRA and the 13,975-weight edit were each trained once. Under the Astra protocol we use two samples per problem rather than four, and test only MATH-500. We chose the strength of our edit in the combination from three values (1, 0.5, 0.25) on MATH-500 itself, with no held-out set; both weaker strengths beat the Astra edit by at least 7.7 points ($p\le10^{-8}$), so the gain does not hinge on one setting, but the 62.2\% figure is the best of three. We did not test combinations with our smaller edits at reduced strength, or other benchmarks at 32k. The bias test covers the four-column edit, on 38 teacher-forced traces and 96 sampled problems, with a different inference engine (Hugging Face rather than vLLM); we have not tested whether the parts of our gradient edits outside the four channels act as constant vectors too. Packing the vectors into one column failed, and we do not know exactly why per-layer placement matters. Random positions were tested only at 1,398 and 139,750 weights. The 50- and 500-example runs drew from a pool made in 57.6 A100-minutes, and we did not measure the cost of generating 50 examples directly. Except at the 8k budget, the edits make thinking mode roughly as good as thinking off. On code they do not help and may slightly hurt (HumanEval with thinking off: 34.1\% to 29.3\%, $p=0.15$). The whole project, including unrelated and failed runs, used about \$49 of cloud compute: about \$30 for the main study and \$19 for the Astra-protocol runs. Some planned runs (edits of 50 and 14 weights, scaled one-column edits) stopped when the first \$30 of credit ran out.

\section{Conclusion}

A weight edit that takes a small model's thinking-mode math score from 15\% to 40--45\% does not need a frontier model. Plain gradient descent finds one in minutes, from 50 of the model's own answers, using as few as 140 weights. The edit stops the loops rather than adding skill. In this bias-free model, its core is a constant vector written through the most constant channels of the first three layers.

The frontier model's edit does something ours does not: given 32k tokens, it reasons better, and it beats our edits by 5--6 points. But it needs those tokens, and below 16k our edits win by large margins. The two edits are not rivals. Added together, with ours at half strength, they reach 62.2\%, 10 points above the Astra edit and 15 above ours, with half the Astra edit's tokens. Weight edits found in different ways can be stacked, and the best result here came from a frontier model's edit and a gradient edit together. Claims about either should be measured against the other, at more than one token budget, with random-position controls and the thinking-off score reported alongside.

\section*{Acknowledgments}

Experiment code and drafting were assisted by Claude (Anthropic), an AI model. The author designed the study, checked all results and takes full responsibility for the content. All experiments ran on rented cloud GPUs (Modal) at a total cost of about \$49.

\end{document}